\documentclass[runningheads]{llncs}

\usepackage{eccv}

\usepackage{eccvabbrv}

\usepackage{graphicx}
\usepackage{booktabs}
\usepackage{tabularx}
\usepackage{colortbl}
\usepackage{adjustbox}
\usepackage{makecell}
\usepackage{multirow}
\usepackage{array}
\usepackage{amsmath}
\usepackage{pifont}
\usepackage{subcaption} 

\usepackage[accsupp]{axessibility}  
\definecolor{lightgray}{rgb}{0.8,0.8,0.8}
\definecolor{darkyellow}{rgb}{0.8,0.6,0.1}
\definecolor{graycolor}{rgb}{0.95,0.95,0.95}
\newcommand{\venue}[1]{{\color{lightgray}\footnotesize #1}}
\newcommand{\gou}[0]{{\ding{52}}}
\newcommand{\cha}[0]{{\ding{55}}}

\usepackage{hyperref}

\usepackage{orcidlink}

\begin{document}

\title{Learning Semantic-Robust Change Detection \\ via Semantic-Invariant Self-Distillation}

\titlerunning{SCDistill}

\author{Jiuhe Qu\inst{1}\orcidlink{0009-0005-9540-1353}\textsuperscript{*} \and
Yingping Liang\inst{1}\orcidlink{0000-0001-5385-0015}\textsuperscript{*} \and
Ying Fu\inst{1}\orcidlink{0000-0002-6677-694X}\textsuperscript{\dag}}

\authorrunning{J.~Qu et al.}

\institute{Beijing Institute of Technology, Beijing, China\\
\email{\{qujiuhe,liangyingping,fuying\}@bit.edu.cn}}

\maketitle

\makeatletter
\begingroup
\renewcommand{\thefootnote}{\fnsymbol{footnote}}
\footnotetext[1]{Equal contribution. \quad \textsuperscript{\dag} Corresponding author.}
\endgroup
\makeatother

\begin{abstract}
Change detection aims to identify semantic changes between remote sensing images. However, features from models are easily disturbed by non-semantic variations, such as illumination, shadows, and atmospheric changes, leading to false alarms and limited generalization in real-world scenarios. In this paper, we propose \textbf{SCDistill}, a framework for learning semantic-robust change detection via semantic-invariant self-distillation. First, to strengthen semantic consistency, we introduce a semantic-invariant self-distillation strategy that learns semantic robustness from perturbed yet semantically consistent data, empowering the change detector to extract disturbance-resistant features and achieve more reliable and accurate semantic change identification. Second, to expand paired data with non-semantic variations, we design a diffusion-based perturbation simulation pipeline that synthesizes complex environmental changes, enabling the model to explicitly learn to distinguish semantic changes from appearance-level fluctuations and reduce false alarms caused by non-semantic disturbances. These components promote robustness from data and representation perspectives, leading to synergistic performance gains. Extensive experiments demonstrate that SCDistill achieves state-of-the-art performance on multiple semantic change detection benchmarks and exhibits strong generalization to binary change detection and change captioning tasks. Code is accessible at \url{https://github.com/elecreak/SCDistill}.
\end{abstract} 

\section{Introduction}

Change detection (CD) plays a crucial role in remote sensing data analysis, aiming to identify semantic changes between bi-temporal images. This capability facilitates numerous applications, including development monitoring \cite{liu2020building}, land-cover mapping \cite{shi2020land, khan2017forest}, and environmental surveillance \cite{gao2019sea}. With the development of deep learning~\cite{TianYe2023CJE,liang2025relation,ZhangTao2024CJE,zhang2026enhancing}, recent works~\cite{change3d, chen2024changemamba, chang2024triple, li2024decoder} have achieved accuracy gains by modeling semantic segmentation and temporal differencing.

Despite this progress, deep learning-based CD remains vulnerable to non-semantic variations—such as illumination shifts, shadows, atmospheric effects, and seasonal transitions—that frequently occur in real-world remote sensing imagery. These disturbances often change surface appearance without altering underlying semantics, yet existing models \cite{chang2024triple,li2024decoder,ding2022bi} may mistakenly interpret them as semantic changes, leading to false alarms and degraded robustness. A key reason is that most CD pipelines heavily rely on vision encoders pre-trained on single-view classification or segmentation tasks \cite{chang2024triple,li2024decoder,ding2022bi}, where representations are typically appearance-sensitive. When such disturbance-sensitive features are propagated into temporal differencing and detection heads, predictions become unstable under real-world environmental fluctuations.

\begin{figure*}[t]
\centering
\begin{center}
    \includegraphics[width=0.99\textwidth]{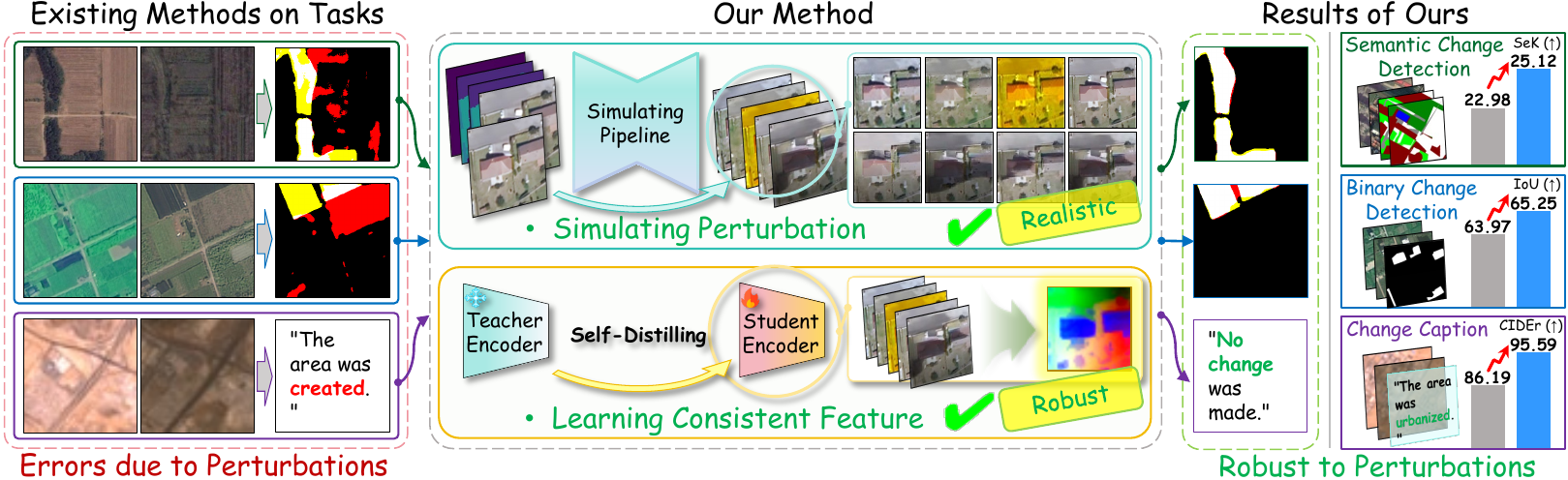}
    \caption{Existing methods rely on limited data collections that lack non-semantic variations and general-purpose encoders that are sensitive to appearance changes. In contrast, \textbf{SCDistill} leverages simulated non-semantic perturbations and a semantic-invariant self-distilled model, achieving consistent improvements across \textbf{three representative tasks}. Regions highlighted in \textcolor{red}{\textbf{Red}} indicate incorrect detections.}
\end{center}
\end{figure*}

To address this issue, we propose \textbf{SCDistill}, a framework for \textbf{semantic-robust change detection} via \textbf{semantic-invariant self-distillation}. The core idea is to explicitly distill semantic invariance into the CD model so that it produces consistent semantic representations and change predictions under diverse appearance conditions. Specifically, we introduce a \textbf{semantic-invariant self-distillation} strategy, where a vision foundation model encoder (VFME) serves as a semantic teacher and guides the student to learn disturbance-resistant representations. Given a clean bi-temporal pair and its appearance-perturbed yet semantically consistent counterpart, the student is trained to align its predictions across conditions to the semantic guidance from teacher. This cross-condition distillation enforces semantic consistency while suppressing appearance-sensitive noise, enabling the student to inherit stronger semantic abstraction and achieve more reliable change detection in complex real-world scenarios.

A practical challenge in applying such semantic-invariant self-distillation is the lack of paired training samples exhibiting semantic-preserving appearance variations. Most existing synthetic CD datasets generate bi-temporal pairs via label-guided synthesis or semantic region replacement, while largely overlooking real-world non-semantic perturbations that occur globally and especially in unchanged regions without label variation \cite{HySCDG,Changen2}. Consequently, models trained on these idealized datasets tend to overfit to clean scenes without non-semantic appearance variations and perfectly aligned conditions, and the distillation objective cannot be sufficiently exercised across diverse environmental conditions.

Motivated by this, from the perspective of data, we design a diffusion-based \textbf{perturbation simulation} pipeline to enrich synthetic CD datasets with realistic, semantic-invariant environmental variations, including illumination shifts, atmospheric and weather changes. This provides the necessary cross-condition supervision for semantic-invariant distillation, allowing the model to explicitly learn to distinguish true semantic changes from appearance-level variations and thereby reduce false alarms. Together, semantic-invariant self-distillation and diffusion-based perturbation simulation jointly improve robustness from representation and data perspectives, leading to more stable and accurate semantic change detection under real-world disturbances.

Extensive experiments demonstrate that SCDistill achieves state-of-the-art performance on multiple semantic change detection benchmarks and generalizes effectively to related tasks, including binary change detection and change captioning, showcasing its strong robustness in real-world scenarios and generalization applicability in related tasks. The proposed components promote robustness from data and representation perspectives, leading to synergistic performance gains. In summary, our main contributions are as follows:
\begin{itemize}
\item We introduce \textbf{SCDistill}, a framework for semantic-robust change detection that mitigates non-semantic disturbances through diffusion-based perturbation simulation and semantic-invariant self-distillation.

\item We design a \textbf{semantic-invariant self-distillation} strategy that learns semantic robustness from perturbed yet semantic-invariant data by self-distillation, enabling disturbance-resistant feature representations for accurate semantic change identification.

\item We propose a diffusion-based \textbf{perturbation simulation} pipeline that synthesizes complex environmental variations, enabling the model to explicitly learn to distinguish semantic changes from appearance-level fluctuations.
\end{itemize}

\section{Related Work}

\noindent\textbf{Change Detection Methods.} CNN-based methods are fundamental to deep visual tasks~\cite{Li2024lhs, zhang2025unaligned, zhang2026real, gcy2026faa} and have been widely used in change detection, with existing designs mainly categorized by bitemporal image integration strategies. Input-level concatenation, such as stacking and differencing~\cite{vorovencii2014change}, or principal component analysis~\cite{abdi2010principal} of bitemporal images before feeding them into a single encoder-decoder, such as Fully Convolutional Network (FCN)~\cite{long2015fully} and U-Net~\cite{ronneberger2015u}, for change prediction. Dual-branch encoders with shared weights process each temporal image, and features are fused by concatenation and subtraction in intermediate layers~\cite{li2023lightweight, chen2022siamese, feng2023change, lei2023ultralightweight}. Other works extend the dual-branch net by incorporating attention mechanisms, such as channel~\cite{hu2018squeeze,wang2020eca}, spatial~\cite{almahairi2016dynamic}, or spatio-temporal~\cite{woo2018cbam, luo2018urban}, to refine feature fusion and emphasize change-relevant regions. However, they remain sensitive to non-semantic disturbances such as illumination and atmospheric changes, often leading to false detections.

\noindent\textbf{Synthetic Datasets for Change Detection.} Synthetic datasets are widely used to alleviate annotation scarcity~\cite{Li2025SFIN, Li2026SCGN, TriMSOD, liang2025Flow, liang2026Lift3Dreamer, zhang2026supervise}, and have recently been used in change detection. Early graphics-based approaches, such as AICD~\cite{bourdis2011constrained} and SyntheWorld~\cite{song2024syntheworld} render 3D scenes but suffer from limited photorealism and diversity. Hybrid pipelines, such as IAug~\cite{chen2021adversarial} and Changen~\cite{ScalableMuti} insert segmented objects into real images using GANs. Diffusion-driven methods like Changen2~\cite{Changen2}, HySCDG~\cite{HySCDG}, and ChangeDiff~\cite{ChangeDiff} synthesize change sequences via generation or text-guided inpainting. These synthetic datasets emphasize semantic transitions but overlook real-world variations like illumination and shadow changes in unchanged regions, causing unreliable supervision.

\noindent\textbf{Vision Foundation Model Encoders.} Deep learning models pre-trained on large-scale datasets serve as vision fundamental feature extract models for downstream vision tasks. Widely adopted architectures include ResNet \cite{he2016deep} trained on ImageNet \cite{deng2009imagenet} and fully convolutional networks (FCN) \cite{long2015fully} for dense prediction. Recently, self-supervised Vision Foundation Model like DINOv3~\cite{simeoni2025dinov3}, composed of vision transformers, has emerged as a powerful alternative, capturing rich semantic representations without manual annotations. Although vision foundation models provide rich semantic priors, their features often lack sufficient discrimination between semantic and appearance-level variations, which can lead to ambiguous representations and unreliable guidance for downstream decoding.

\section{Method}

\begin{figure*}[t]
\centering
\includegraphics[width=1.0\textwidth]{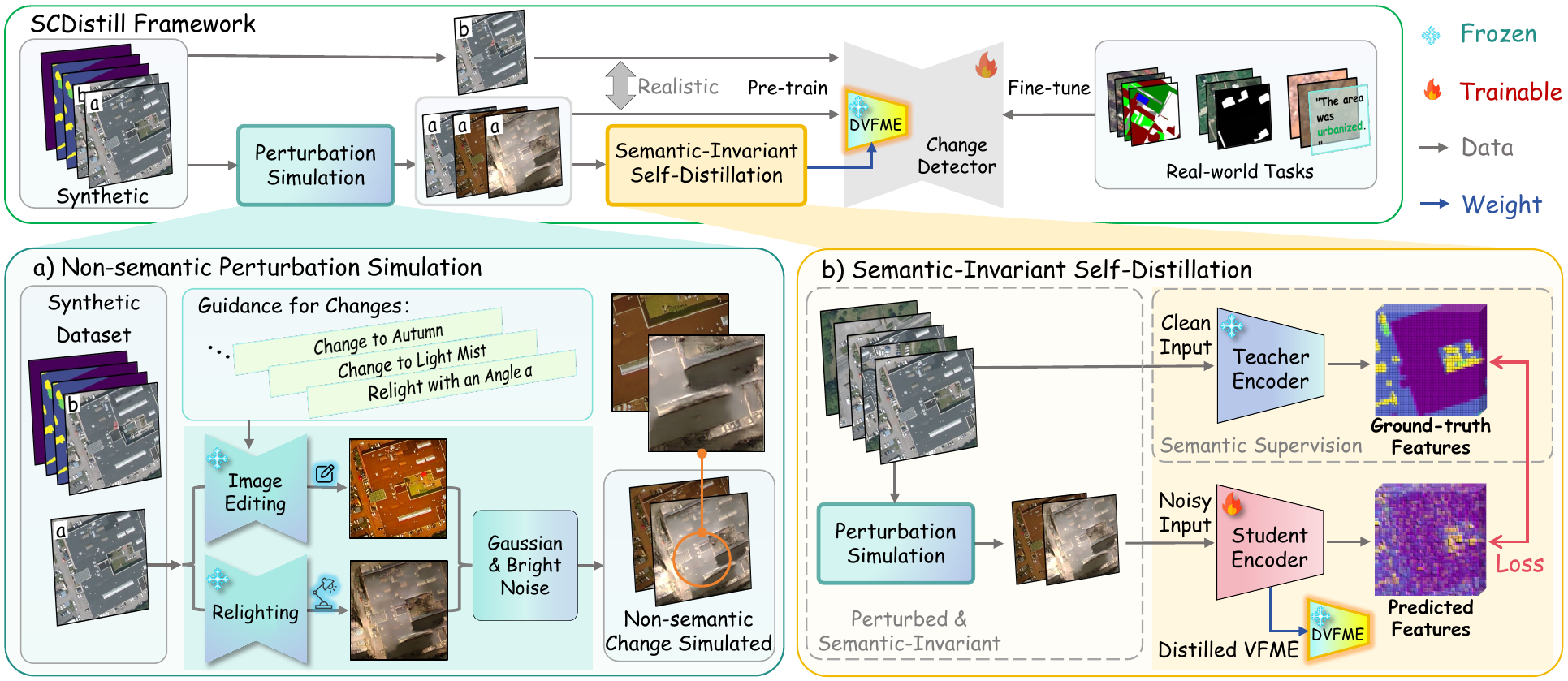}
\caption{Overview of the proposed SCDistill framework. a) The diffusion-based perturbation simulation synthesizes environmental variations to explicitly model real-world appearance disturbances, enhancing robustness against non-semantic changes. b) Meanwhile, the vision foundation model encoder obtain semantic supervision from the clean image in semantic-invariant self-distillation, guiding the disturbed student to learn disturbance-resistant representations for accurate semantic change detection.}
\label{fig_main}
\end{figure*}

In this section, we first introduce the formulation and motivation of our approach. Then, we present the perturbation simulation, where a pipeline synthesizes non-semantic disturbances to model realistic appearance variations. Next, we describe the proposed semantic-invariant self-distillation strategy that learns disturbance-resistant semantic representations for robust semantic change detection, as illustrated in Figure~\ref{fig_main}. Finally, we provide the learning details.

\subsection{Formulation and Motivation}

Change detection (CD) aims to identify differences between bi-temporal images $\{\mathbf{I}_1, \mathbf{I}_2\}$ of the same scene. Depending on the output form, CD can be performed at different levels, including semantic CD, binary CD, and change captioning. All these tasks share a common encoder for feature $\mathbf{F}$ extraction, with task-specific decoders producing change maps or textual descriptions $\mathbf{C}$:
\begin{equation}
\begin{aligned}
\mathbf{F} = \text{Enc}(\{\mathbf{I}_1, \mathbf{I}_2\}), \quad 
\mathbf{C} = \text{Dec}(\mathbf{F}).
\end{aligned}
\end{equation}
Since large-scale bi-temporal datasets with pixel-level annotations are difficult to collect, some recent works synthesize data by generating semantic layouts $\mathbf{S}$ and producing corresponding images through guided rendering:
\begin{equation}
\begin{aligned}
\{\mathbf{I}_1, \mathbf{I}_2, \mathbf{L}\} = \text{Synthesize}(\mathbf{S}; \Phi),
\end{aligned}
\end{equation}
where $\Phi$ denotes the synthesis process that generates bi-temporal images $\{\mathbf{I}_1, \mathbf{I}_2\}$ and their semantic labels $\mathbf{L}$.

However, there are still two main challenges. First, existing synthetic datasets primarily emphasize semantic region changes, while neglecting realistic non-semantic variations in unchanged regions, such as illumination shifts, shadows, and atmospheric effects, commonly observed in real-world imagery. Models trained on such data tend to overfit to idealized semantic transitions and struggle to distinguish true semantic changes from appearance-level disturbances. To address this, we propose a \textbf{perturbation simulation} strategy that explicitly models non-semantic variations using a diffusion-based renderer, converting existing synthetic datasets into more realistic bi-temporal scenes. This exposes the model to physically plausible disturbances during training, fostering more robust visual reasoning under real-world conditions.

Moreover, even with realistic perturbations, standard visual encoders pretrained on single-view tasks remain sensitive to appearance fluctuations, producing inconsistent semantic representations that undermine downstream CD. We introduce a \textbf{semantic-invariant self-distillation} framework that transfers high-level semantic knowledge from a clean-image teacher to a student trained on perturbed counterparts. This cross-condition distillation enforces semantic consistency while filtering out non-semantic noise, yielding disturbance-robust features that enable reliable and accurate change identification. Together with perturbation simulation, this approach addresses both data and feature level limitations, forming a framework for semantic-robust change detection.

\subsection{Non-semantic Perturbation Simulation}

Synthetic change detection datasets alleviate the scarcity of annotated data, yet they usually fail to simulate the non-semantic changes that are widely present in real scenarios. Consequently, models trained on synthetic data often show poor robustness to such disturbances, which adversely affects their fine-tuning performance. To address this limitation, we augment synthetic datasets with complex environmental non-semantic variations that imitate realistic variations between two temporally captured images of the same scene. Specifically, we analyze real-world bi-temporal change detection datasets and identify typical non-semantic change types that frequently lead to false detections in existing methods. These include changes in illumination and shadow caused by sunlight direction and weather variations, which significantly influence model predictions.

\noindent\textbf{Data Source and Notation.}
To enable the simulated non-semantic perturbation samples to be learned by the CD model, we start from an existing synthetic change detection dataset that provides semantically edited image pairs. This dataset is generated by modifying specific regions of real remote sensing images to create semantic changes. Given a sample triplet $\{\mathbf{X}_0, \mathbf{X}_1, \mathbf{L}\}$, where $\mathbf{X}_0$ and $\mathbf{X}_1$ are pre- and post-change images and $\mathbf{L}$ is the semantic change labels, since non-semantic disturbances between bi-temporal images can be introduced by altering only one image of each pair, we apply these perturbations to the pre-change image $\mathbf{X}_0$, which can be formulated as:
\begin{equation}
\begin{aligned}
\tilde{\mathbf{X}}_0 = \text{Perturb}(\mathbf{X}_0), \quad \tilde{\mathbf{X}}_1 = \mathbf{X}_1, \quad \tilde{\mathbf{L}} = \mathbf{L}.
\end{aligned}
\end{equation}
This preserves the semantic editing applied in the synthetic dataset while injecting realistic appearance variations between the bi-temporal image pairs.

\noindent\textbf{Non-semantic Variations Simulation.}
Different categories of non-semantic changes are simulated using different models. For illumination and shadow variations induced by solar angle changes, we employ a scene illumination modification model such as DIR \cite{yang2025relighting} to simulate lighting conditions. For weather and atmospheric variations, we adopt an image editing model such as InstructPix2Pix \cite{brooks2023instructpix2pix}, guided by multiple pre-defined textual prompts $p$ describing various non-semantic changes. Through these processes, multiple perturbations simulated versions of $\mathbf{X}_0$ are produced, which can be formulated as:
\begin{equation}
\begin{aligned}
\tilde{\mathbf{X}}_0^{(i)} = \{\text{Relighting}(\mathbf{X}_0), \text{Editing}(\mathbf{X}_0, p_i)\}, \quad i = 1,\dots,N.
\end{aligned}
\end{equation}
This results in a set of noise-injected images $\{ \tilde{\mathbf{X}}_0^{(1)}, \tilde{\mathbf{X}}_0^{(2)}, \dots \}$ that reflect a wide range of realistic non-semantic changes. Combined with the original $\mathbf{X}_1$ and label $\mathbf{L}$, this forms a new large-scale dataset that simulates both semantic and non-semantic changes, bridging the gap between synthetic and real-world data.

Compared with GAN-based~\cite{gan-based-edit} methods or only making modifications at the pixel level, such as Gaussian blur, random noise, and color jitter, the diffusion-based simulation offers controllable and diverse perturbations with high visual fidelity, ensuring the generated non-semantic variations remain consistent with real-world conditions. This ensures the dataset bridges the domain gap and preserves semantic integrity crucial for subsequent distillation.

\subsection{Semantic-Invariant Self-Distillation}

While the perturbation simulation exposes the model to realistic non-semantic variations, it does not directly guarantee semantic robustness at the feature level. Therefore, we introduce a semantic-invariant self-distillation strategy to explicitly inject such robustness into the encoder representations. Most existing semantic change detection methods employ general-purpose visual encoders pretrained on single-image recognition tasks, such as the ResNet~\cite{he2016deep} used in Bi-SRNet~\cite{ding2022bi} or the X3D~\cite{feichtenhofer2020x3d} backbone in Change3D~\cite{change3d}. However, these encoders, trained on single-view datasets, are insufficient to capture semantic consistency across complex real-world variations, resulting in degraded robustness in semantic feature extraction. To address this limitation, we introduce a semantic-invariant encoder that learns to extract noise-resistant representations by distilling knowledge from diverse non-semantic variations synthesized from large-scale data.

\noindent\textbf{Noise-Augmented Pairs Construction.}
To enable this distillation, we prepare a dataset consisting of image pairs with simulated non-semantic perturbations while maintaining semantic invariance. For each pair, the first image is sampled from a single-temporal remote sensing dataset, and the second is generated by applying the non-semantic variation simulation pipeline described in perturbation simulation, which means the perturbation-simulated dataset can be used as this dataset. Additionally, random low-level perturbations such as Gaussian blur, grayscale conversion, and brightness and saturation shifts are applied to further increase pixel-level non-semantic discrepancies. As a result, each image pair shares identical semantic content but differs in appearance due to realistic non-semantic variations.

\noindent\textbf{Robust Representations Learning.}
We employ a distillation-based training strategy to transfer semantic invariance into the pre-trained vision foundation model encoder (VFME). Given a single-temporal image $\mathbf{X}$, we first generate its noise-augmented version $\tilde{\mathbf{X}}=\text{Perturb}(\mathbf{X})$. We use the VFME pretrained on general tasks as a frozen teacher and to initialize the student. The teacher $\text{E}_T$ extracts a clean semantic representation $\mathbf{F}_T$ from the original image, which serves as the target for the student model $\text{E}_s$ trained on the noisy input affected by non-semantic perturbations, which can be formulated as:
\begin{equation}
\begin{aligned}
\mathbf{F}_T = \text{E}_T(\mathbf{X}), \quad  \mathbf{F}_S = \text{E}_S(\tilde{\mathbf{X}}),
\end{aligned}
\end{equation}
where $\mathbf{F}_S$ and $\mathbf{F}_T$ are features from student and teacher. We then minimize the discrepancy between $\mathbf{F}_S$ and $\mathbf{F}_T$ using L2 regression loss in distilling.

Through training on diverse noise conditions, the distilled VFME, as the student encoder, learns to produce semantic-invariant semantic features that remain consistent with clean outputs from the teacher model. Importantly, this training requires no change labels or semantic segmentation supervision, allowing it to scale across large unlabeled datasets. The resulting encoder is then used in downstream semantic change detection models to enhance robustness against real-world appearance shifts. Through this process, the student encoder internalizes semantic invariance, serving as a robust foundation for subsequent change detection models under diverse environmental conditions. The teacher only receives clean inputs and provides semantic-invariant supervision as targets of distillation. So the robustness of the student is not inherited from a robust teacher, but is mainly induced by the semantic-invariant distillation process.

\subsection{Learning Details} 
To implement the perturbation simulation, we use the DIR~\cite{yang2025relighting} model to simulate illumination changes, where the azimuth angle is uniformly from 0° to 360° and the elevation angle from 20° to 70°. When employing InstructPix2Pix~\cite{brooks2023instructpix2pix} for weather and atmospheric editing, a randomly selected pre-defined prompt is used for each input image to direct the modification process.

To construct a change detector guided by feature representations with robustness against non-semantic interference, we adopt the video encoder–universal decoder from Change3D~\cite{change3d} and integrate the disturbance-robust semantic encoder distilled from the VFME into the encoder branch. This self-distilled encoder provides semantically stable features $\mathbf{F}_0$ and $\mathbf{F}_1$ for the pre-change and post-change images $(\mathbf{I}_0, \mathbf{I}_1)$, which are fused with the shallow X3D~\cite{feichtenhofer2020x3d} representations to enhance resilience against non-semantic perturbations and guide the deeper layers to focus on meaningful semantic change regions while suppressing irrelevant appearance-level fluctuations.

During the training of our CD network, the model is pre-trained on the diffusion-augmented synthetic dataset to learn invariance and fine-tuned on real-world benchmarks. Following~\cite{change3d}, we adopt task-specific joint losses as follows cross-entropy, Dice~\cite{dice}, and cosine similarity~\cite{ding2022bi} for semantic CD; cross-entropy and Dice for binary CD; and cross-entropy for change captioning.

\begin{table}[t]
\small
    \centering
    \caption{Performance comparison of training on multiple semantic CD benchmarks with the state-of-the-art method. The best results are represented in bold, and the second-best results are indicated with an underline. All results are described aspercentages (\%).}
    \renewcommand{\arraystretch}{1.12} 
    \resizebox{\textwidth}{!}{
    \begin{tabular}{l l c c c c c c c c }
        \toprule
        \multirow{2}{*}{\textbf{Method}} & \multirow{2}{*}{\textbf{Backbone}} & \multicolumn{4}{c}{\textbf{SECOND~\cite{yang2021asymmetric}}} & \multicolumn{4}{c}{\textbf{HRSCD~\cite{daudt2019multitask}}} \\ \cmidrule(lr){3-6} \cmidrule(lr){7-10}
         & & Fscd & mIoU & OA & SeK & Fscd & mIoU & OA & SeK \\  \midrule
HRSCD-S4~\cite{daudt2019multitask} \venue{[CVIU'2019]} & ResNet & 58.21 & 71.15 & 86.62 & 18.80 & 69.39 & 67.79 & 81.32 & 22.50 \\
SCDNet~\cite{peng2021scdnet} \venue{[JPRS'2022]} & ResNet & 60.01 & 70.97 & 87.40 & 19.73 & 70.80 & 67.43 & 81.46 & 23.61 \\
ChangeMask~\cite{zheng2022changemask} \venue{[IJAEOG'2021]} & EfficientNet-B0 & 59.74 & 71.46 & 86.93 & 19.50 & 70.59 & 67.56 & 81.65 & 23.43\\
SSCD-L~\cite{ding2022bi} \venue{[TGRS'2022]} & ResNet & 61.22 & 72.60 & 87.19 & 21.86 & 69.69 & 66.21 & 81.05 & 21.88  \\
Bi-SRNet~\cite{ding2022bi} \venue{[TGRS'2022]} & ResNet & 61.85 & 72.08 & 87.20 & 21.36 & 71.72 & 67.83 & 82.06 & 24.59\\
MTSCD~\cite{cui2023mtscd} \venue{[IJAEOG'2023]} & ResNet & 60.23 & 71.68 & 87.04 & 20.57 & 67.56 & 63.71 & 72.06 & 9.03\\
JFRNet~\cite{chang2024triple} \venue{[TGRS'2024]} & ResNet & 62.63 & 72.82 & 87.10 & 22.56 & 66.65 & 64.65 & 76.20 & 18.78\\
DEFO~\cite{li2024decoder} \venue{[TGRS'2024]} & ResNet & 61.18 & 72.39 & 87.01 & 21.08 & 66.49 & 65.67 & 81.36 & 19.20 \\
ChangeMamba~\cite{chen2024changemamba} \venue{[TGRS'2024]} & Mamba-based  & 61.59 &	72.32 &	\underline{87.84} &	20.30  & 70.04 & 67.93 &	82.68 &	24.50  \\ 
Changen2~\cite{Changen2} \venue{[TPAMI'2024]} & ViT & 62.79 & 72.39 & 87.67 & 22.09 & 73.03 & \underline{68.73} & \underline{82.82} & 26.25 \\ %
Change3D~\cite{change3d} \venue{[CVPR'2025]} & VideoEncoder & \underline{62.83} & \underline{72.95} & 87.42 & \underline{22.98} & \underline{73.29} & 68.67 & 82.57 & \underline{26.85}\\
\midrule
\rowcolor{graycolor} SCDistill (Ours) & Distilled VFME & \textbf{65.64} & \textbf{74.01} & \textbf{88.59} & \textbf{25.12} & \textbf{74.29} & \textbf{70.28} & \textbf{83.61} & \textbf{28.38} \\  
        \bottomrule 
    \end{tabular}
    }
    \label{table_SCD}
\end{table}

\begin{figure*}[t]
\centering
\includegraphics[width=1.0\textwidth]{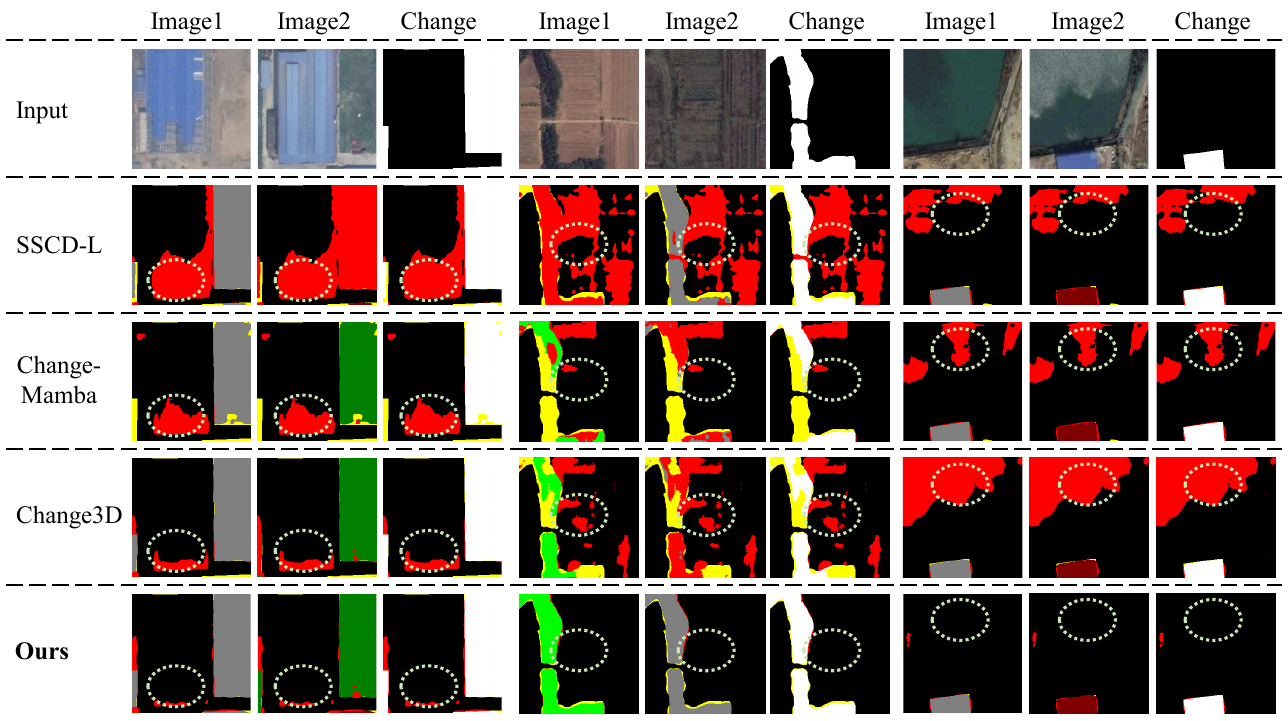} 
\caption{Qualitative comparison with representative state-of-the-art methods on the SECOND~\cite{yang2021asymmetric} dataset. Black represents non-change, \textcolor{red}{\textbf{Red}} and \textcolor{darkyellow}{\textbf{Yellow}} indicate \textcolor{red}{\textbf{False}} and \textcolor{darkyellow}{\textbf{Missed}} detections, while white and other colors denote correct change regions and their semantic categories respectively.}
\label{fig_vis_main}
\end{figure*}

\begin{table}[t]
\small  \centering 
    \caption{Performance comparison of training on binary CD task benchmarks.}
    \setlength{\tabcolsep}{3pt}
    \renewcommand{\arraystretch}{1.05} 
    \resizebox{0.98\columnwidth}{!}{
        
    \begin{tabular}{l c c c c c c c c c }
        \toprule
        \multirow{2}{*}{\textbf{Method}} & \multicolumn{3}{c}{\textbf{CLCD~\cite{liu2022cnn}}} & \multicolumn{3}{c}{\textbf{LEVIR-CD~\cite{levir-cd}}} & \multicolumn{3}{c}{\textbf{WHU-CD~\cite{whu-cd}}} \\ \cmidrule(lr){2-4} \cmidrule(lr){5-7} \cmidrule(lr){8-10}
          & F1 & IoU & OA & F1 & IoU & OA & F1 & IoU & OA \\  
        \midrule
DTCDSCN~\cite{liu2020building} \venue{[GRSL'2020]}  
& 57.47 & 40.81 & 94.59
& 87.43 & 77.67 & 98.75 
& 79.92 & 66.56 & 98.05 \\
SNUNet~\cite{fang2021snunet} \venue{[GRSL'2021]}  
& 60.82 & 43.63 & 94.90
& 88.16 & 78.83 & 98.82 
& 83.22 & 71.26 & 98.44 \\
ChangeFormer~\cite{bandara2022transformer} \venue{[IGRSS'2022]}  
& 61.31 & 44.29 & 94.98
& 90.40 & 82.48 & 99.04 
& 87.39 & 77.61 & 99.11 \\
BIT~\cite{chen2021remote} \venue{[TGRS'2021]} 
& 59.93 & 42.12 & 94.77 
& 89.31 & 80.68 & 98.92 
& 83.98 & 72.39 & 98.52 \\
ICIFNet~\cite{feng2022icif} \venue{[TGRS'2022]}  
& 68.66 & 52.27 & 95.77
& 89.96 & 81.75 & 98.99 
& 88.32 & 79.24 & 98.96 \\
Changer~\cite{fang2023changer} \venue{[TGRS'2023]}  
& 67.07 & 50.46 & 95.69 
& 90.70 & 82.99 & 99.08
& 89.18 & 80.48 & 99.17 \\
DMINet~\cite{feng2023change} \venue{[TGRS'2023]}  
& 67.24 & 50.65 & 95.21
& 90.71 & 82.99 & 99.07 
& 88.69 & 79.68 & 98.97 \\
GASNet~\cite{zhang2023global} \venue{[ISPRS'2023]}  
& 63.84 & 46.89 & 94.01
& 90.52 & 83.48 & 99.07 
& 91.75 & 84.76 & 99.34 \\
AMTNet~\cite{liu2023attention} \venue{[ISPRS'2023]}  
& 75.10 & 60.12 & 96.45
& 90.76 & 83.08 & 98.96 
& 92.27 & 85.64 & 99.32 \\
Self-Pair~\cite{selfpair} \venue{[WACV'2023]}    
& 75.49 & 60.63 & 96.42
& 90.83 & 83.20 & 99.07 
& 91.59 & 84.94 & 99.03 \\  
Changen~\cite{zheng2023changen} \venue{[ICCV'2023]}  
& 76.96 & 62.88 & 96.31      
& 91.12 & 83.69 & 99.11 
& 92.47 & 87.10 & 98.95 \\ 
EATDer~\cite{ma2023eatder} \venue{[TGRS'2024]}  
& 72.01 & 56.19 & 96.11
& 91.20 & 83.80 & 98.75 
& 90.01 & 81.97 & 98.58 \\ 
AnyChange~\cite{zheng2024anychange} \venue{[NeurIPS'2024]}  
& 75.93 & 61.20 & 96.51 
& 90.65 & 82.91 & 99.06 
& 92.87 & 86.02 & 99.32 \\ 
Changen2~\cite{Changen2} \venue{[TPAMI'2024]}   
& 78.00 & 63.84 & \underline{96.89}  
& 91.79 & 84.62 & \underline{99.16} 
& \underline{94.62} & \underline{89.75} & 99.43 \\  
Change3D~\cite{change3d} \venue{[CVPR'2025]}    
& \underline{78.03} & \underline{63.97} & 96.87
& \underline{91.82} & \underline{84.87} & \textbf{99.17} 
& 94.56 & 89.69 & \underline{99.57}  \\ \midrule
\rowcolor{graycolor}
\textbf{SCDistill (Ours)}      
& \textbf{78.97} & \textbf{65.25} & \textbf{96.93}
& \textbf{91.88} & \textbf{85.04} & \underline{99.16} 
& \textbf{94.79} & \textbf{89.84} & \textbf{99.60} \\ 
        \bottomrule 
    \end{tabular}
}
\label{table_BCD}
\end{table}

\section{Experiments}
In this section, we first introduce the datasets and evaluation metrics for experiments. Then, performance comparisons are conducted with the other methods. Finally, ablations, visualized qualitative results, and discussions are provided to confirm the effectiveness of the main components. 

\subsection{Experimental Settings}

\noindent\textbf{Datasets.} To provide abundant data for pretraining and to cover a wide range of semantic change types, we adopt the large-scale synthetic dataset FSC-180k\cite{HySCDG}, which includes 20 categories of semantic change simulated from real remote sensing images across diverse land-cover types. For evaluation, multiple real-world benchmarks are utilized, including HRSCD~\cite{daudt2019multitask} and SECOND~\cite{yang2021asymmetric}, which contain diverse non-semantic variations such as illumination, shadow, and atmospheric changes commonly observed in real scenarios. For binary CD and change captioning tasks, we further employ CLCD~\cite{liu2022cnn} and DUBAI-CC~\cite{hoxha2022change}, as their feature are diverse land-cover transitions and higher semantic complexity, posing greater challenges to model robustness and generalization. During both training and testing, all datasets are processed to a spatial resolution of $256\times256$. Specifically, for CD datasets, bi-temporal image pairs and corresponding labels are synchronously cropped into non-overlapping patches, while for change captioning datasets, image resolutions are directly resized to the same scale.

\noindent\textbf{Evaluation Metrics.} For semantic CD, we report Fscd (F1 score on changed regions), mean Intersection over Union (mIoU), Overall Accuracy (OA), and Separate Kappa (SeK). For binary CD, we use standard metrics including pixel-wise F1 score, IoU, and Overall Accuracy (OA). For change captioning, we use BLEU-1 and BLEU-2 for their focus on lexical and phrasal precision, which is important for CD tasks, along with METEOR, ROUGE-L, and CIDEr-D, as adopted in change captioning literature~\cite{liu2022remote}. 

\noindent\textbf{Training Parameters.} For the self-distillation of the semantic encoder, we employ the AdamW optimizer with a learning rate of $2\times10^{-5}$ and a weight decay of $0.05$, using a batch size of eight per GPU. For the pretraining and fine-tuning of the change detection model, we adopt the Adam optimizer with $\beta=(0.9, 0.99)$, a weight decay of $2\times10^{-4}$, and an initial learning rate of $2\times10^{-4}$ that decays following the schedule $(1 - \text{iter}_{\text{curr}} / \text{iter}_{\text{max}})^{\alpha} \times \text{lr}$, where $\alpha$ is set to $0.9$. All models are trained on $256\times256$ image patches with data augmentation techniques including random resizing, flipping, and cropping. Distillation takes 5 hours, pretraining takes 20 hours as a one-time cost. The simulation pipeline takes only 8 hours on 1 NVIDIA RTX 3090, with 0.7 seconds per image. The fine-tuning deployment takes 8 hours, as the pretrained model and self-distilled encoder trained by the perturbation simulated data can be used across tasks directly without cost.

\subsection{Main Results}

\noindent\textbf{Quantitative Results on Semantic CD.} As shown in Table~\ref{table_SCD}, we compare our method with representative semantic CD approaches under standard supervised settings, including classical models (HRSCD-S4~\cite{daudt2019multitask}, SCDNet~\cite{peng2021scdnet}), recent convolution-based frameworks (SSCD-L~\cite{ding2022bi}, Bi-SRNet~\cite{ding2022bi}, DEFO~\cite{li2024decoder}, MTSCD~\cite{cui2023mtscd}, JFRNet~\cite{chang2024triple}), and advanced designs such as ChangeMask~\cite{zheng2022changemask} (EfficientNet-B0 backbone), ChangeMamba~\cite{chen2024changemamba} (Mamba-based), Changen2~\cite{Changen2} (pretrained with generative change supervision), and Change3D~\cite{change3d} (X3D-based video encoder~\cite{feichtenhofer2020x3d}). In contrast to these methods, our method integrates the semantic-invariant self-distillation with non-semantic perturbations simulation, enabling the model to suppress non-semantic feature noise and maintain semantic consistency. This design yields consistent improvements across benchmarks, with particularly notable gains in semantic-related metrics; for example, on SECOND, our method surpasses the state-of-the-art method by 9.31\% in SeK and 4.47\% in Fscd, confirming its robustness under complex scenarios.

\noindent\textbf{Qualitative Results on Semantic CD.} Figure~\ref{fig_vis_main} presents qualitative comparisons between our method and three representative baselines, selected to cover diverse architectural paradigms: SSCD-L~\cite{ding2022bi} (ResNet-based), ChangeMamba~\cite{chen2024changemamba} (Mamba-based), and Change3D~\cite{change3d} (video-encoder-based, current state-of-the-art). These models represent different backbone families and performance tiers, providing a comprehensive visual comparison of semantic change detection quality. Our method achieves more accurate change localization and semantic classification in complex real-world scenes, while other methods tend to misclassify areas affected by non-semantic environmental variations such as shadows, illumination, or weather shifts. These results visually confirm that SCDistill effectively suppresses irrelevant appearance changes and maintains semantic consistency, enabling robust deployment in challenging environments.

\noindent\textbf{Generalization to Binary CD.} As shown in Table~\ref{table_BCD}, we evaluate the generalization ability of our SCDistill framework from semantic CD to binary CD under standard supervised settings. We compare our approach with representative methods, including DTCDSCN~\cite{liu2020building}, SNUNet~\cite{fang2021snunet}, ChangeFormer~\cite{bandara2022transformer}, BIT~\cite{chen2021remote}, ICIFNet~\cite{feng2022icif}, Changer~\cite{fang2023changer}, DMINet~\cite{feng2023change}, GASNet~\cite{zhang2023global}, AMTNet~\cite{liu2023attention}, Self-Pair~\cite{selfpair}, Changen~\cite{zheng2023changen}, EATDer~\cite{ma2023eatder}, AnyChange~\cite{zheng2024anychange}, Changen2~\cite{Changen2}, and Change3D~\cite{change3d}. Across all challenging datasets, our method consistently outperforms existing approaches, achieving state-of-the-art performance. In particular, we observe significant gains in accuracy-sensitive metrics (F1 and IoU), demonstrating that the semantic-invariant self-distillation and synthetic perturbation pretraining effectively enhance the robustness to non-semantic variations of the encoder and improve its generalization capability across change detection tasks.

\begin{table}[t]  \centering 
\caption{Performance comparison of training on change captioning task benchmark DUBAI-CC~\cite{hoxha2022change} with state-of-the-art methods.}
    \setlength{\tabcolsep}{3pt}
    \resizebox{0.8\columnwidth}{!}{
\small
\begin{tabular}{l c c c c c c}  \toprule
Method & BLEU-1 & BLEU-2 & METEOR & ROUGE & CIDEr\\  \midrule
DUDA~\cite{park2019robust} \venue{[ICCV'2019]} & 58.82 & 43.59 & 22.05 & 48.34 & 62.78 \\
MCCF-S~\cite{qiu2021describing} \venue{[ICCV'2021]} & 52.97 & 37.02 & 18.64 & 43.29 & 53.81 \\
MCCF-D~\cite{qiu2021describing} \venue{[ICCV'2021]} & 64.65 & 50.45 & 25.09 & 51.27 & 66.51 \\
RSICC~\cite{liu2022remote} \venue{[TGRS'2022]} & 67.92 & 53.61 & 25.41 & 51.96 & 66.54 \\
SEN~\cite{liu2023decoupling} \venue{[TGRS'2024]} & 64.12 & 50.41 & 23.67 & 48.19 & 65.15 \\
PromptCC~\cite{zhou2024single} \venue{[TGRS'2023]} & 70.03 & 58.41 & 26.48 & 55.82 & 85.44 \\
Change3D~\cite{change3d} \venue{[CVPR'2025]} & \underline{72.25} & \underline{58.68} & \underline{27.06} & \underline{56.04} & \underline{86.19} \\ \midrule
\rowcolor{graycolor} SCDistill (Ours) & \textbf{72.39} & \textbf{59.77} & \textbf{29.48} & \textbf{59.65} & \textbf{95.59} \\  \bottomrule \end{tabular}
}
\label{table_CC}
\end{table}

\noindent\textbf{Generalization to Change Captioning.} 
As shown in Table~\ref{table_CC}, we further assess the generalization of our SCDistill framework to the change captioning task. We compare against recent advanced approaches, including DUDA~\cite{park2019robust}, MCCFormer~\cite{qiu2021describing}, RSICC~\cite{liu2022remote}, SEN~\cite{liu2023decoupling}, PromptCC~\cite{zhou2024single}, and Change3D~\cite{change3d}. 
Our method achieves superior performance across all benchmarks, especially on semantically sensitive metrics such as METEOR, ROUGE-L, and CIDEr. These improvements indicate that our semantic-invariant self-distillation strategy enables the model to maintain semantic consistency under various appearance perturbations, thereby producing more precise and semantically faithful textual descriptions of changes.

\subsection{Discussions}
To further understand the effectiveness of our SCDistill framework, we conduct comprehensive ablation studies and visual analyses focusing on the perturbation simulation, the semantic-invariant self-distillation, few-shot experiments evaluating the generalization ability of the components, the feature-level behavior after self-distillation, and the influence of different teachers in self-distillation.

\noindent\textbf{Effect of Perturbation Simulation.} 
To evaluate the contribution of our diffusion-based perturbation simulation, we remove this component and train the model using only the standard synthetic semantic CD datasets. As shown in Table~\ref{ablation_table}, the absence of perturbation simulation causes a significant drop in performance. Besides, the joint design of our components forms a closed-loop mechanism and achieves larger gains. For example, +2.78 Fscd compared to +0.69 or +1.05 individually, exceeding what would be expected from a simple additive effect. This confirms that constructing a synthetic dataset enriched with non-semantic disturbances provides a more realistic and robust pretraining signal, enabling the model to better distinguish semantic transitions from appearance variations and thus enhancing its generalization to real-world conditions.

\noindent\textbf{Effect of Semantic-invariant Self-distillation.} 
We further evaluate the impact of the semantic-invariant self-distillation strategy by comparing the model using the original VFME without distillation against our self-distilled encoder. As reported in Table~\ref{ablation_table}, removing the self-distillation strategy consistently degrades performance. This demonstrates that the proposed semantic-invariant self-distillation effectively transfers semantic robustness from clean to perturbed representations, guiding the encoder to learn disturbance-resistant semantic features and yielding more reliable and accurate semantic change detection.

\noindent\textbf{Effect of the Main Components via Few-shot Generalization.}
We further evaluate different ablated settings under a few-shot protocol on the SECOND~\cite{yang2021asymmetric} benchmark to assess generalization from synthetic to real-world data. Specifically, models pretrained with different component combinations are fine-tuned using limited labeled samples. As shown in Table~\ref{supp_low_table}, each component contributes to improved few-shot performance, while the full model consistently achieves the best results. This demonstrates that both perturbation simulation and semantic-invariant self-distillation enhance transferability, enabling strong generalization from synthetic pretraining data to real-world change detection scenarios.

\begin{table}[t]\centering
\small
    \centering
    \caption{Effect of the main components.}
    \setlength{\tabcolsep}{4pt}
    \resizebox{0.7\columnwidth}{!}{
\begin{tabular}{l | c c | c c c c}\toprule
\textbf{Setting} & Simulate & Distill & Fscd & mIoU & OA & SeK\\  \midrule
Vanilla & \cha{} & \cha{} & 62.86 & 72.54 & 87.53 & 22.30 \\
+ Simulation  & \gou{} & \cha{} & 63.55 & 73.10 & 87.87 & \underline{23.39} \\
+ Distillation  & \cha{} & \gou{} & \underline{63.91} & \underline{73.13} & \underline{88.24} & 23.37 \\ 
\rowcolor{graycolor} Ours & \gou{} & \gou{} & \textbf{65.64} & \textbf{74.01} & \textbf{88.59} & \textbf{25.12}  \\ \bottomrule  
    \end{tabular}
    }
\label{ablation_table}
\end{table}

\begin{table}[t]\centering
\small
    \centering
    \caption{Results of few-shot ablation studies on SECOND~\cite{yang2021asymmetric}.}
    \setlength{\tabcolsep}{3pt}
    \resizebox{0.99\columnwidth}{!}{
\begin{tabular}{l c c c c c c c c c c c c}\toprule
\multirow{2}{*}{\textbf{Setting}}  & \multicolumn{4}{c}{\textbf{1\%}} & \multicolumn{4}{c}{\textbf{10\%}} & \multicolumn{4}{c}{\textbf{30\%}}\\ \cmidrule(lr){2-5} \cmidrule(lr){6-9} \cmidrule(lr){10-13}
 & Fscd & mIoU & OA & SeK & 
 Fscd & mIoU & OA & SeK  & 
 Fscd & mIoU & OA & SeK \\  \midrule
w/o Simulation & 
\underline{51.8}&	\underline{67.6}&	\underline{84.8}&	\underline{11.7}&	
58.9&	\underline{71.3}&	87.3&	\underline{19.1}&
\underline{62.8}&	\underline{72.4}&	87.5&	\underline{22.2}\\
w/o Distillation & 
 48.9 &	67.2 &	84.2 &	9.8 &		
 \underline{59.6} &	70.0 &	\underline{87.9} &	18.5 &		
 61.8 &	72.3 &	\underline{87.9} &	21.4\\
\rowcolor{graycolor} \textbf{Ours} & 
\textbf{55.6}&	\textbf{68.7}&	\textbf{86.0}&	\textbf{15.0}&	
\textbf{62.5}&	\textbf{72.4}&	\textbf{88.0}&	\textbf{21.6}&		
\textbf{64.4}&	\textbf{73.3}&	\textbf{88.4}&	\textbf{23.5}\\ \bottomrule  
    \end{tabular}
    }
\label{supp_low_table}
\end{table}

\noindent\textbf{Effect of Self-distillation via Quantitative and Qualitative Analysis.}
We first provide quantitative evaluation of semantic invariance before and after distillation. Table~\ref{cosine_abla_table} reports MSE, cosine similarity, and centered kernel alignment (CKA) computed on features extracted from unchanged real-world HRSCD~\cite{daudt2019multitask} image pairs by encoders before and after distillation. These metrics measure feature discrepancy (MSE), directional consistency (cosine similarity), and representation-level alignment (CKA), respectively. Results show that the distilled encoder achieves lower MSE and higher cosine similarity and CKA, indicating improved robustness to real-world appearance variations. 

\begin{figure}[t]
\centering
\includegraphics[width=0.98\columnwidth]{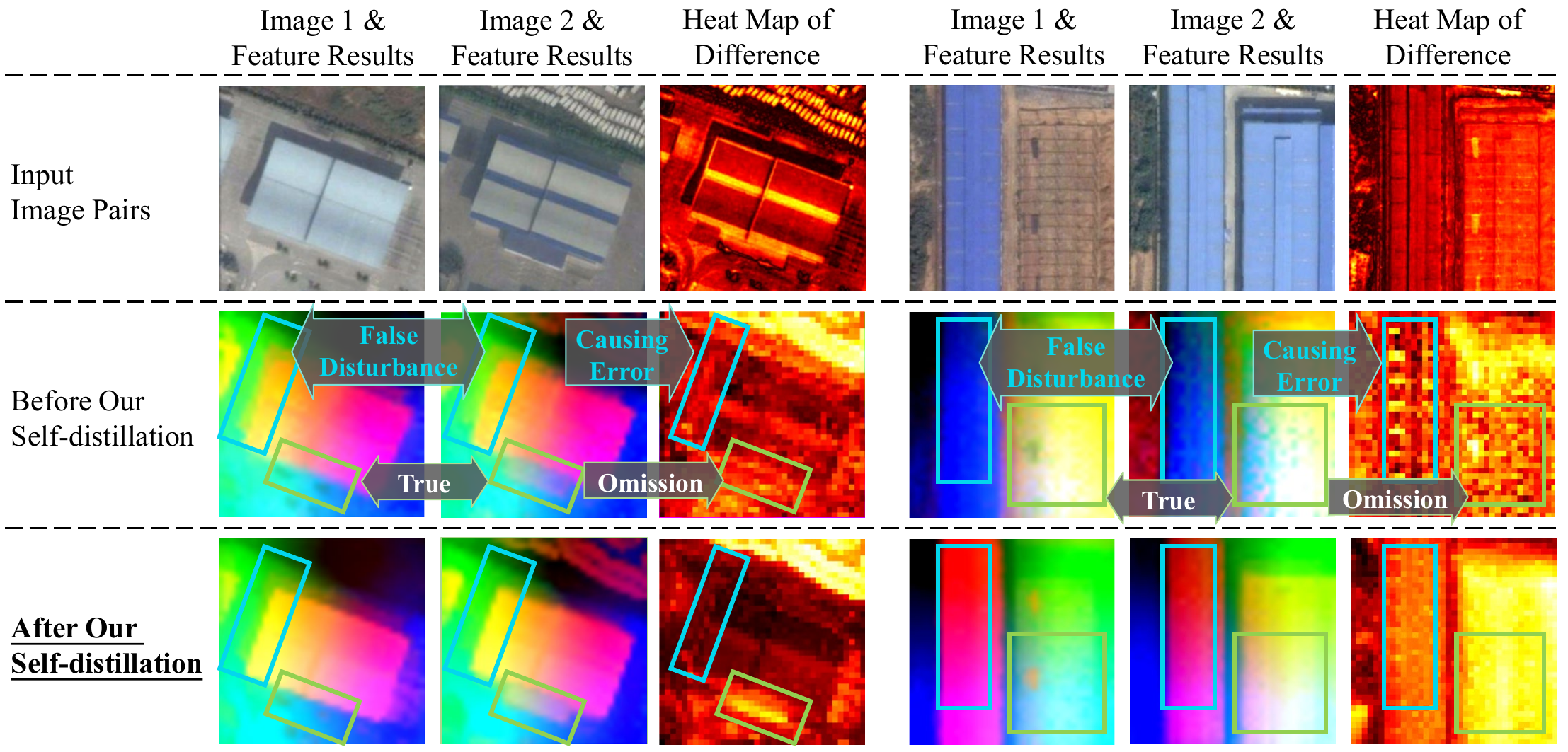} 
\captionof{figure}{Comparison of feature representations via PCA~\cite{pca} before and after semantic-invariant self-distillation on the real-world SECOND dataset~\cite{yang2021asymmetric}.}
\label{figdvt}
\end{figure}

Furthermore, we visualize the feature differences between bi-temporal inputs as heatmaps, comparing the outputs of the teacher and student models (Fig.~\ref{figdvt}). The teacher model shows strong responses to illumination, shadows, and other appearance-level variations, leading to spurious activations and weak separation between semantic and non-semantic changes. In contrast, the self-distilled model produces disturbance-resistant and semantically consistent feature differences, effectively distinguishing true semantic transitions from non-semantic fluctuations and providing more robust guidance for change detection.

\noindent\textbf{Effect of Self-distilling by Different Teachers.} 
We further examine the impact of different vision foundation model encoders (VFMEs), including DINOv2~\cite{oquab2023dinov2} and DINOv3~\cite{simeoni2025dinov3}, with and without our semantic-invariant self-distillation, as shown in Table~\ref{vfm_ablation}. For both teachers, introducing distillation consistently brings substantial performance gains, demonstrating that our semantic-invariant self-distillation effectively improves robustness regardless of the specific teacher model. This indicates that the robustness of the distilled encoder is mainly induced by enforcing semantic consistency across perturbed inputs during distillation, rather than being directly inherited from a robust teacher. Among the teachers, DINOv3 achieves the best overall results, likely due to its stronger semantic representations on clean natural images, which provide more reliable semantic supervision for our distillation process.

\begin{figure}[t]
    \begin{minipage}[b]{0.4\textwidth}
    \centering
    \captionof{table}{Quantitative effect of self-distillation. (\%)}
    \label{cosine_abla_table}
    \setlength{\tabcolsep}{5pt}
    \renewcommand{\arraystretch}{1.2} 
    \resizebox{0.98\columnwidth}{!}{
\begin{tabular}{l c c c}\toprule
\textbf{Distilling} & MSE$\downarrow$ & cosine$\uparrow$ & CKA$\uparrow$ \\  \midrule
Before & 6.41 & 84.05 & 87.08 \\
\rowcolor{graycolor} \textbf{After} & \textbf{3.32} & \textbf{89.78} & \textbf{90.89} \\ \bottomrule
    \end{tabular}
    }
    \end{minipage}
    \hfill
    \begin{minipage}[b]{0.52\textwidth}

\captionof{table}{Comparison of different teachers w and w/o semantic-invariant self-distillation.}
\label{vfm_ablation}
    \centering
    \renewcommand{\arraystretch}{1.1} 
    \setlength{\tabcolsep}{5pt}
    \resizebox{0.98\columnwidth}{!}{
    \begin{tabular}{l c c c c}\toprule
\textbf{Teacher model} & Fscd & mIoU & OA & SeK\\  \midrule
DINOv2~\cite{oquab2023dinov2} & 63.09 & 73.00 & 87.92 & 22.80 \\
\rowcolor{graycolor} + Our Distillation  & 
\underline{64.28} & \underline{73.42} & \underline{88.41} & \underline{23.81}\\ 
 DINOv3~\cite{simeoni2025dinov3} & 63.55 & 73.10 & 87.87 & 23.39 \\
\rowcolor{graycolor} + Our Distillation &  \textbf{65.64} & \textbf{74.01} & \textbf{88.59} & \textbf{25.12} \\ \bottomrule  
    \end{tabular}
    }

    \end{minipage}

\end{figure}

\section{Conclusion} 
In this paper, we introduce \textbf{SCDistill}, a framework that enables semantic-robust change detection via semantic-invariant self-distillation. Our approach tackles the limitations of conventional methods, which often mistakenly detect non-semantic variations, such as illumination, shadows, and atmospheric changes, as true semantic changes, leading to degraded robustness in real-world scenarios. Specifically, SCDistill introduces a semantic-invariant self-distillation strategy to transfer semantic robustness from perturbed yet semantic-preserving data, guiding the encoder to learn disturbance-resistant and semantically stable representations. In addition, our method incorporates a diffusion-based perturbation simulation pipeline that synthesizes diverse non-semantic environmental disturbances, allowing the model to explicitly learn to distinguish semantic changes from appearance-level fluctuations. Extensive experiments on multiple benchmarks demonstrate that our SCDistill achieves state-of-the-art performance and strong generalization across various change detection tasks.


\section*{Acknowledgements}
This work is supported by the National Natural Science
Foundation of China (62331006), the Fundamental Research Funds for the Central Universities, and the National Natural Science Foundation of China under Grant (625B2026).

%
%
\bibliographystyle{splncs04}
\bibliography{main}
\end{document}